\documentclass[acmsmall,screen,nonacm]{acmart}

\setcopyright{none}
\renewcommand\footnotetextcopyrightpermission[1]{}

\usepackage{graphicx}   
\usepackage{subcaption}   
\usepackage{tikz}
\usetikzlibrary{arrows.meta,positioning}

\definecolor{tilebg}{rgb}{0.87,0.94,1}
\usepackage{tcolorbox}

\tcbset{
  smtsimple/.style={
    colback=white,
    colframe=black,
    boxrule=0.7pt,
    top=5pt, bottom=5pt,
    boxsep=0pt,
  }
}

\newcommand{\QFLIA}{\texttt{QF\char`_ LIA}}
\newcommand{\kind}[1]{\mathsf{#1}}

\title{From Patterns to Maze Structures: SMT-Based Path Synthesis and 2D/3D Construction}

\author{Shengyi Wang}
\affiliation{%
  \institution{Shanghai Qi Zhi Institute}
  \city{Shanghai}
  \country{China}}
\email{wangshengyi@sqz.ac.cn}

\date{}     

\begin{document}

\begin{abstract}

  We present a pipeline for constructing maze structures from input
  patterns such as text or shapes. The central path-synthesis problem
  is encoded in Satisfiability Modulo Theories as global constraints
  on adjacency, continuity, and pattern-constrained coverage, allowing
  each fixed-bound instance to be solved in one call. The resulting
  path is either a planar, self-avoiding route or a layered traversal
  with prescribed over--under crossings, and it serves as a scaffold for
  constructing planar mazes and three-dimensional realizations of woven
  mazes. This report extends the published Bridges 2026 conference
  paper with more representative SMT-LIB examples and a fuller account
  of how synthesized paths become concrete maze constructions in
  planar and three-dimensional form.

\end{abstract}

\maketitle

\section{Introduction}

This report extends the Bridges 2026 conference paper
\cite{Wang2026PatternsMazeSolutions}, which introduced the SMT
formulation for synthesizing maze solution paths from input patterns.
The present version keeps that formulation but broadens the scope from
path synthesis to maze construction. In particular, it gives more
representative SMT-LIB fragments, describes how synthesized paths are
completed into planar and woven maze structures, and develops the
height information needed to realize layered crossings as
three-dimensional maze geometry.

\begin{figure}[htbp]
  \centering
  \begin{minipage}[b]{0.4\textwidth}
    \includegraphics[width=\textwidth]{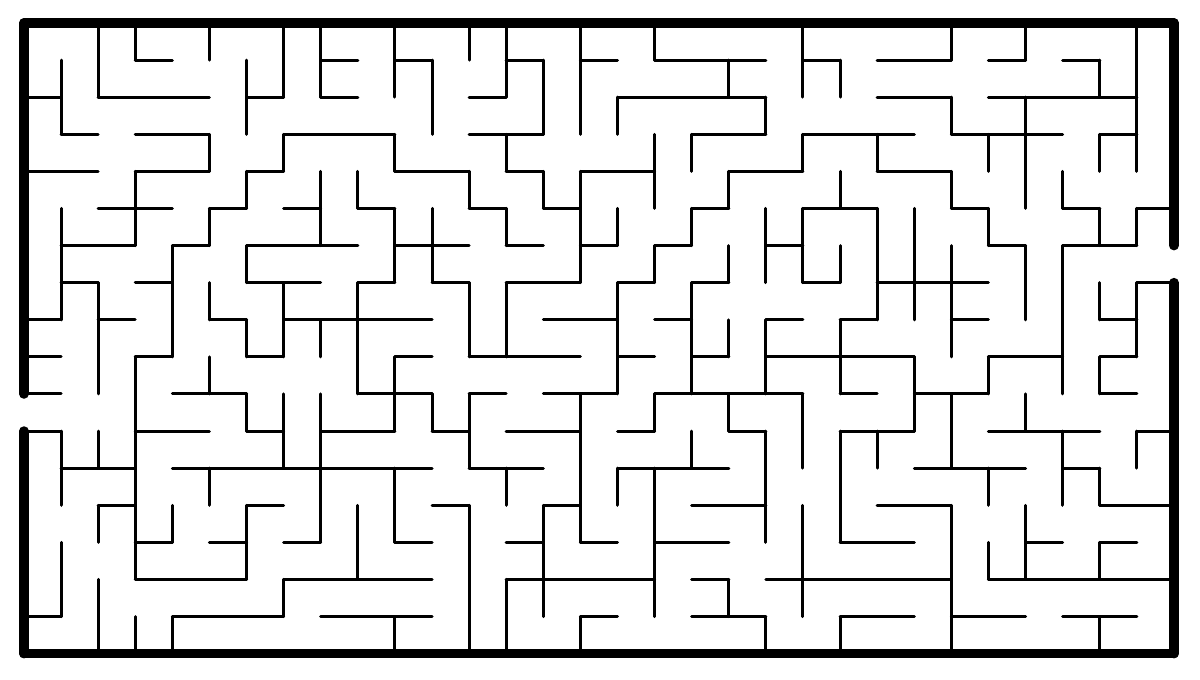}
    \subcaption{}
    \label{fig:1a}
  \end{minipage}
  ~
  \begin{minipage}[b]{0.4\textwidth}
    \includegraphics[width=\textwidth]{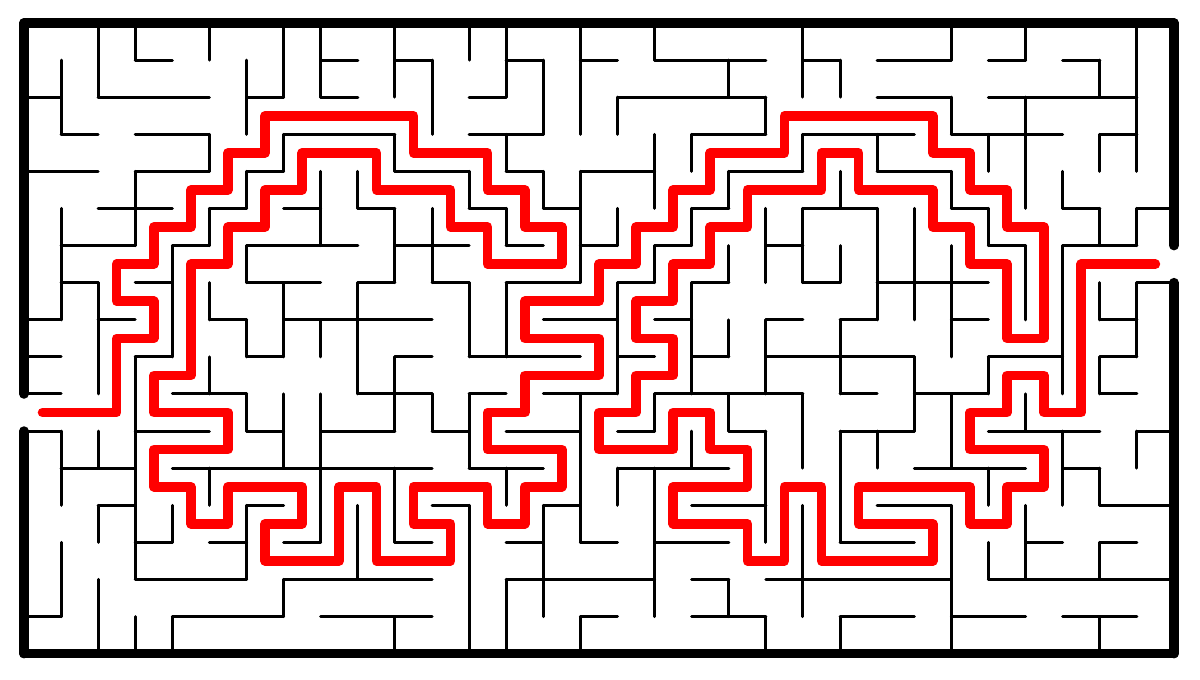}
    \subcaption{}
    \label{fig:1b}
  \end{minipage}
  \\
  \begin{minipage}[b]{0.4\textwidth}
    \includegraphics[width=\textwidth]{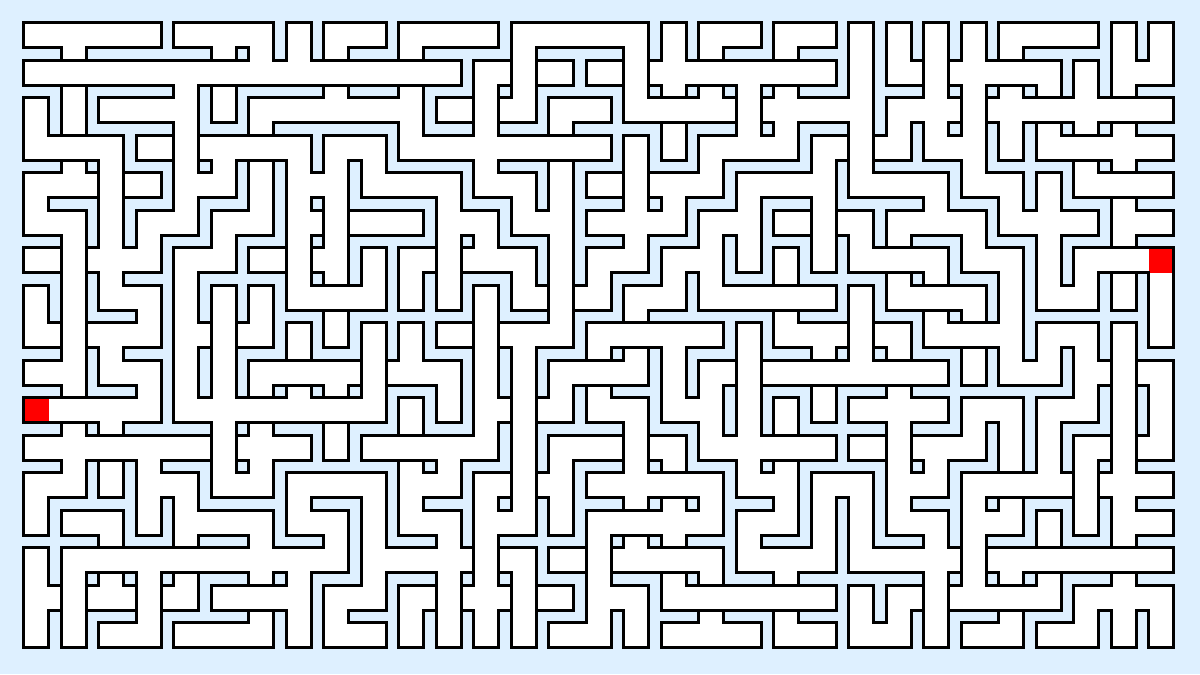}
    \subcaption{}
    \label{fig:1c}
  \end{minipage}
  ~
  \begin{minipage}[b]{0.4\textwidth}
    \includegraphics[width=\textwidth]{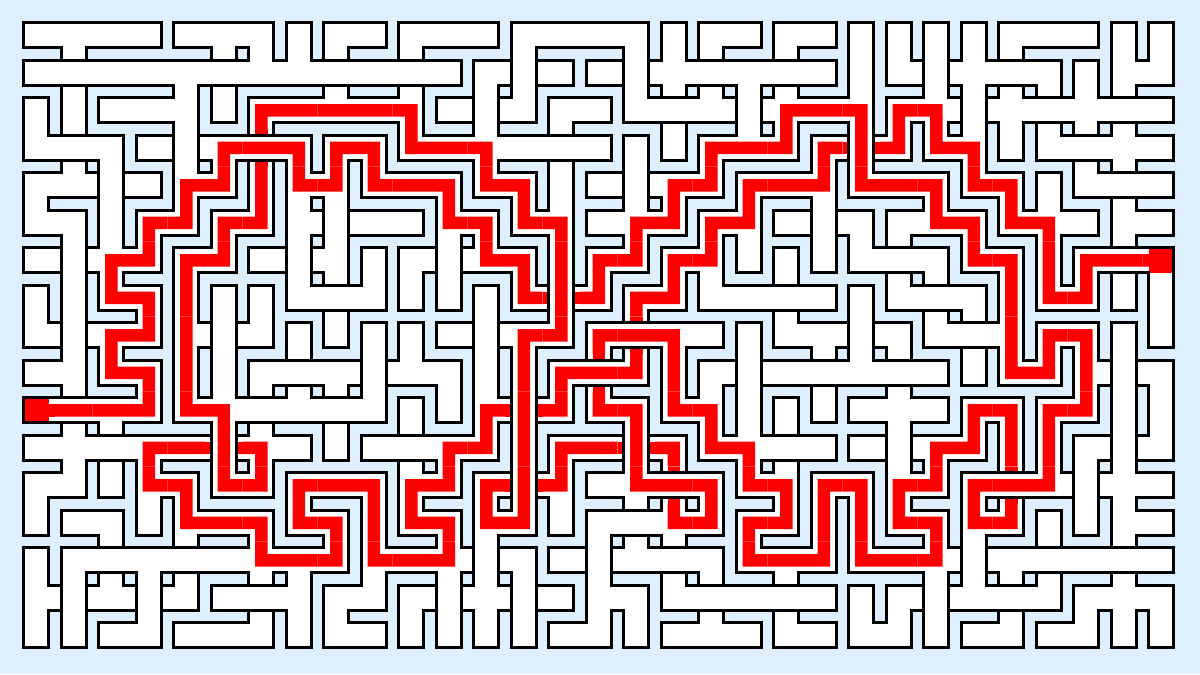}
    \subcaption{}
    \label{fig:1d}
  \end{minipage}
  \caption{Planar and woven mazes derived from the same ``$\infty$''
    solution pattern: (a) Planar maze, \\ (b) Planar maze with solution,
    (c) Woven maze, (d) Woven maze with solution.}
  \Description{Four maze images derived from an infinity-shaped pattern, showing planar and woven versions with and without highlighted solutions.}
  \label{fig:inf}
\end{figure}

We derive two distinct maze solution paths from the same input
pattern: a planar path and a layered path with crossings. Each
solution path then serves as a scaffold for maze construction.
Figure~\ref{fig:inf} gives an example of the resulting planar and
woven mazes, both with and without highlighted solution paths.

\begin{figure}[htbp]
  \centering
  \resizebox{0.96\linewidth}{!}{%
    \begin{tikzpicture}[
  box/.style={
    draw=black!60,
    rounded corners=3pt,
    fill=tilebg,
    inner xsep=6pt,
    inner ysep=6pt,
    align=center
  },
  arr/.style={
    -{Latex[length=3mm,width=2mm]},
    line width=1.5pt,
    draw=black!70
  },
  lab/.style={font=\small, fill=white, inner sep=1pt}
]

  \node[box] (p) {Input Pattern\\(text or image)};
  \node[box, right=16mm of p] (g) {Grid Positions\\(rasterized)};
  \node[box, right=16mm of g] (s) {Solution Path\\(planar or layered)};
  \node[box, right=16mm of s] (m) {Maze\\Construction};

  \draw[arr] (p) -- node[lab, above] {Step 1} (g);
  \draw[arr] (g) -- node[lab, above] {Step 2} (s);
  \draw[arr] (s) -- node[lab, above] {Step 3} (m);

\end{tikzpicture}%
  }
  \caption{Three-step pipeline from an input pattern to a maze.}
  \Description{A left-to-right pipeline from input pattern to grid positions to solution path to maze construction.}
  \label{fig:pipeline}
\end{figure}
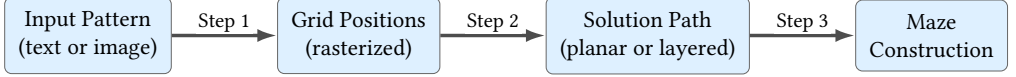

The construction of the mazes considered in this paper can be viewed
as a three-step process, as shown in Figure~\ref{fig:pipeline}. First,
a given text or image is converted into a set of grid
positions. Second, a path is constructed on this grid, starting from a
prescribed entry position and ending at a prescribed exit position,
while traversing as many target grid positions as possible. Depending
on the intended maze type, this path may be required to be planar or
may admit layered crossings. Third, a maze is generated from the
resulting path.

Among these steps, the first is straightforward and can be handled
using existing image processing tools, such as rasterization followed
by binary thresholding to extract a target set of grid positions from
the input pattern. The conference version focused primarily on the
second step: finding a suitable path that satisfies global structural
constraints while remaining faithful to the input pattern. This report
retains that formulation and develops the third step in greater
detail, including both planar maze completion and a
three-dimensional realization of layered crossings.

The third step builds on standard maze-generation ideas, for example
those surveyed in~\cite{Buck2015MazesForProgrammers}. From a
graph-theoretic perspective, planar maze generation can be viewed as
constructing a spanning tree of an underlying grid graph subject to
the synthesized solution path. In the layered case, the crossing
relation supplies additional topological information that must be
turned into compatible height data before a three-dimensional maze
geometry can be constructed.

\section{Path Synthesis}

This section describes the synthesis of the solution path that serves
as the combinatorial scaffold for the later maze construction. The
path is modeled as a sequence of moves on grid positions, with
additional crossing information when woven paths are allowed. At this
stage, the goal is not yet to construct maze walls or three-dimensional
geometry, but to obtain a valid path satisfying the prescribed local
movement rules, global connectivity requirements, and coverage
objective. We first compare this task with more familiar path
optimization problems, then describe the SMT encoding, and finally
summarize the observed solver performance.

\subsection{Problem Analysis}

We first consider the planar setting shown in Figures~\ref{fig:1a}
and~\ref{fig:1b}, where the maze solution is an orthogonal,
self-avoiding path on a grid. Given a set of grid positions derived
from an input pattern, the task is to construct a path from a
prescribed entry position to a prescribed exit position that visits as
many of these positions as possible.

A natural first comparison is with the Traveling Salesman Problem
(TSP), an NP-complete problem~\cite{Papadimitriou1977EuclidTSP}. In
its classical formulation, TSP seeks a shortest path visiting a given
set of positions, and such shortest solutions typically avoid
self-intersections. This makes it tempting to apply existing TSP
solvers to the present setting~\cite{Cook2012InPursuit}. In an ideal
case, if a path visits \(n\) grid positions by successive moves
between adjacent positions, its total length is exactly \(n-1\). Any
diagonal segment would have length \(\sqrt{2}>1\) and therefore cannot
appear in a truly optimal solution when a unit-length step between
neighboring grid positions is available.

In practice, however, this ideal behavior cannot be relied upon for
two reasons. First, the set of grid positions induced by an input
pattern may not admit a path that achieves the \(n-1\) bound, so even
an optimal solution may necessarily include longer
connections. Second, on large instances, practical solvers may return
non-optimal paths, in which diagonal shortcuts appear. In both cases,
the resulting solutions violate the intended orthogonal,
grid-following structure and tend to require substantial ad hoc
post-processing.

From a graph-theoretic perspective, the problem is more accurately
described as a Hamiltonian path problem on a grid graph, rather than a
metric optimization problem. Hamiltonian path problems are NP-complete
in general graphs~\cite{GareyJohnson1979}, and this hardness persists
even for grid graphs~\cite{hamiltonGrid}, ruling out simple
constructive or purely local solutions for arbitrary patterns.

Existing SAT encodings for Hamiltonian path/cycle already require
nontrivial global connectivity
constraints~\cite{Zhou2020EfficientSAT}, and extending them to
optimization variants is correspondingly more cumbersome.

An alternative line of work is suggested by tile-based methods for
constructing single-line drawings, such as the approach of Bosch and
Snyder~\cite{BoschSnyder2023TileBasedSingleLine}. In that setting,
local path fragments are assembled subject to global consistency
constraints, and the overall problem is formulated as (mixed-)integer
linear program. The initial solution may contain additional disjoint
loops, which can be handled either by merging them or by adding
loop-elimination inequalities (in the spirit of subtour elimination)
and re-solving the model.

In our work, we adopt the same tile-based representation of local path
fragments, but instead of formulating the overall problem as an
integer program, we cast the global connectivity and pattern coverage
constraints in a Satisfiability Modulo Theories (SMT) encoding
tailored to maze paths. Modern SMT solvers provide a natural framework
for expressing both propositional structure and integer constraints,
allowing each fixed-bound existence question to be decided in one
solver call.

One major advantage of the tile-based representation is its ability to
express structurally distinct path types within a single unified
framework. Under this representation, the planar and layered cases
differ only in the acceptability of crossing tiles: planar paths
forbid crossings, while layered paths allow them. This illustrates
that both path structures arise from the same local primitives; the
planar case is obtained simply by disallowing crossing tiles.

\subsection{SMT Encoding}

Before describing the constraint encoding, we first formalize the
underlying path construction problem. Let \(P \subset \mathbb{N}^2\)
be the finite set of grid positions \((r,c)\) (row/column indices)
obtained by rasterizing the input pattern and retaining the positions
corresponding to foreground pixels. Two positions \((r,c),(r',c') \in
\mathbb{N}^2\) are said to be adjacent if and only if
\(|r-r'|+|c-c'|=1\). Given two designated positions \(s,t \in P\), the
objective is to find an orthogonal path
\[
s = a_0, a_1, \ldots, a_n = t,
\]
where each \(a_i \in P\), consecutive positions are adjacent. Grid
positions are visited at most once in the planar setting and at most
twice in the layered setting. Among all such paths, we seek one that
maximizes the number of visited positions, i.e., the length \(n\).

\begin{figure}[h!tbp]
  \centering
  \resizebox{0.96\linewidth}{!}{%
    \begin{tikzpicture}[
  tile/.style={
    draw=gray!60,
    fill=tilebg,
    line width=0.5pt
  },
  pathseg/.style={
    draw=black,
    line width=6pt,
    line cap=butt
  },
  num/.style={
    anchor=north
  }
]

\def\s{1.6}    
\def\gap{0.5} 
\pgfmathsetmacro{\w}{0.5*\s} 

\newcommand{\tiletwo}[3]{%
  \begin{scope}[shift={({(#1)*(\s+\gap)},0)}]
    \draw[tile] (0,0) rectangle (\s,\s);
    #2
    \node[num] at (\w,-0.05*\s) {#3};
  \end{scope}
}

\tiletwo{0}{\draw[pathseg] (\w,\s) -- (\w,\w) -- (\s,\w);}{1}

\tiletwo{1}{\draw[pathseg] (\w,0) -- (\w,\w) -- (\s,\w);}{2}

\tiletwo{2}{\draw[pathseg] (0,\w) -- (\w,\w) -- (\w,0);}{3}

\tiletwo{3}{\draw[pathseg] (0,\w) -- (\w,\w) -- (\w,\s);}{4}

\tiletwo{4}{\draw[pathseg] (\w,0) -- (\w,\s);}{5}

\tiletwo{5}{\draw[pathseg] (0,\w) -- (\s,\w);}{6}

\tiletwo{6}{
  \draw[pathseg] (\w,0) -- (\w,\s);
  \draw[pathseg] (0,\w) -- (\s,\w);
}{7}

\tiletwo{7}{}{8}

\end{tikzpicture}%
  }
  \caption{Tile types encoding local path connectivity patterns.}
  \Description{Eight tile types, each representing a local path connectivity pattern or an empty tile.}
  \label{fig:tiles}
\end{figure}
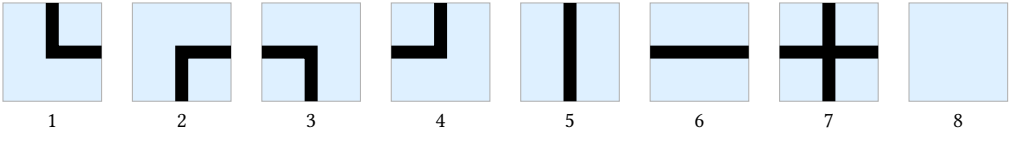

To encode this path construction problem, we first represent paths
using a tile-based formulation. Each grid position is associated with
a tile specifying a local path connectivity pattern. The tile types
shown in Figure~\ref{fig:tiles} serve as the basic primitives of this
representation. Under this formulation, the existence of a valid path
is reduced to the problem of placing tiles so that local connectivity
constraints are satisfied and global path structure emerges.

We then encode the resulting tile placement problem in the
quantifier-free theory of linear integer arithmetic (\QFLIA), a
fragment of SMT that permits reasoning about Boolean conditions
together with integer linear constraints. In this encoding, Boolean
variables represent adjacency and ordering decisions induced by tile
connectivity, while integer constraints enforce global properties such
as connectivity, degree, and coverage. For a fixed coverage bound, one
call to an SMT solver determines whether a valid path satisfying all
constraints exists, returning a model when the instance is satisfiable
and reporting unsatisfiability otherwise.

For each grid position \((r,c) \in P\), we introduce an integer-valued
tile variable \(t_{r,c} \in \{1,2,\dots,8\}\), whose value selects one
of the tile types shown in Figure~\ref{fig:tiles}. Local connectivity
semantics are specified by a collection of auxiliary functions on tile
types, summarized in~\eqref{eq:tile-defs}. These definitions form the
basis for all subsequent local and global constraints. Each tile
variable \(t_{r,c}\) is declared as an integer constrained by
\(\mathrm{inRange}\).
\begin{equation}
\label{eq:tile-defs}
\setlength{\jot}{2pt}
\begin{aligned}
\mathrm{inRange}(t)
&= 1 \le t \le 8,
&
\mathrm{hasUp}(t)
&= t \in \{1,4,5,7\},
&
\mathrm{hasRight}(t)
&= t \in \{1,2,6,7\},
\\
\mathrm{nonWhite}(t)
&= \mathbf{1}_{t \neq 8},
&
\mathrm{hasDown}(t)
&= t \in \{2,3,5,7\},
&
\mathrm{hasLeft}(t)
&= t \in \{3,4,6,7\}.
\end{aligned}
\end{equation}
These definitions can be encoded in SMT-LIB, a standard input language
for SMT solvers \cite{SMTLIBStandard}. For example:
\begin{tcolorbox}[smtsimple]
\begin{verbatim}
(define-fun hasUp ((t Int)) Bool (or (= t 1) (= t 4) (= t 5) (= t 7)))
(define-fun nonWhite ((t Int)) Int (ite (= t 8) 0 1))
\end{verbatim}
\end{tcolorbox}

Using the indicator function \(\mathrm{nonWhite}\) defined
in~\eqref{eq:tile-defs}, we measure the length of a candidate path by
the total number of grid positions that participate in it. Concretely,
we define the coverage value
\begin{equation}
\label{eq:coverage}
C = \sum_{(r,c)\in P} \mathrm{nonWhite}(t_{r,c}).
\end{equation}
Depending on solver support, this coverage can either be maximized
directly or constrained to exceed a prescribed lower bound. In
SMT-LIB, the latter is expressed by asserting a coverage constraint:
\begin{tcolorbox}[smtsimple]
\begin{verbatim}
(assert (>= (+ (nonWhite tile_3_1) (nonWhite tile_3_2) ...) K))
\end{verbatim}
\end{tcolorbox}
When supported, this constraint can alternatively be replaced by a
direct maximization objective.

\subsubsection{Local Connectivity Constraints}
Local connectivity constraints enforce consistency between neighboring
tiles and regulate how the path interacts with the boundary of the
grid. We first impose adjacency consistency, and then handle boundary
behavior, distinguishing general boundary positions from the
designated entry and exit.

For any two horizontally adjacent grid positions \((r,c)\) and
\((r,c+1)\), a path segment exiting to the right of \((r,c)\) must
enter from the left of \((r,c+1)\). An analogous condition holds for
vertical adjacency. Formally, for all adjacent pairs of grid positions
within the domain, we require
\[
\begin{aligned}
\mathrm{hasRight}(t_{r,c}) &\;\Longleftrightarrow\;
\mathrm{hasLeft}(t_{r,c+1}), \\
\mathrm{hasDown}(t_{r,c}) &\;\Longleftrightarrow\;
\mathrm{hasUp}(t_{r+1,c}).
\end{aligned}
\]

At the boundary of the induced domain \(P\), path segments are not
allowed to leave \(P\): whenever a neighboring position is absent from
\(P\), that direction must be disabled. For any position \((r,c)\in
P\) other than the designated entry and exit positions, we impose the
following boundary-closure constraints:
\[
\begin{aligned}
(r,c)\notin\{s,t\},\ (r,c-1)\notin P
&\;\Rightarrow\; \neg \mathrm{hasLeft}(t_{r,c}), \\
(r,c)\notin\{s,t\},\ (r,c+1)\notin P
&\;\Rightarrow\; \neg \mathrm{hasRight}(t_{r,c}), \\
(r,c)\notin\{s,t\},\ (r-1,c)\notin P
&\;\Rightarrow\; \neg \mathrm{hasUp}(t_{r,c}), \\
(r,c)\notin\{s,t\},\ (r+1,c)\notin P
&\;\Rightarrow\; \neg \mathrm{hasDown}(t_{r,c}).
\end{aligned}
\]

The designated entry and exit positions \(s=(r_s,c_s)\) and
\(t=(r_t,c_t)\) are assumed to lie on the boundary and are treated as
explicit exceptions to the boundary constraints. To ensure a single
connection to the exterior, the tile type at each endpoint is fixed
according to its boundary location: if an endpoint lies on the left or
right boundary, a horizontal tile is enforced; if it lies on the top
or bottom boundary, a vertical tile is enforced. Concretely, we impose
\[
\begin{aligned}
(r_s,c_s-1)\notin P \ \vee\ (r_s,c_s+1)\notin P
&\;\Rightarrow\; t_{r_s,c_s}=6, &
(r_s-1,c_s)\notin P \ \vee\ (r_s+1,c_s)\notin P
&\;\Rightarrow\; t_{r_s,c_s}=5, \\
(r_t,c_t-1)\notin P \ \vee\ (r_t,c_t+1)\notin P
&\;\Rightarrow\; t_{r_t,c_t}=6, &
(r_t-1,c_t)\notin P \ \vee\ (r_t+1,c_t)\notin P
&\;\Rightarrow\; t_{r_t,c_t}=5.
\end{aligned}
\]
where tile types \(5\) and \(6\) correspond to vertical and horizontal
path segments, respectively. If an endpoint happens to lack both a
horizontal and a vertical neighbor in \(P\), we break the tie by
prioritizing the horizontal direction, i.e., we fix the endpoint tile
to be horizontal.

The above constraints translate directly into SMT-LIB assertions. Here
are a few representative examples:
\begin{tcolorbox}[smtsimple]
\begin{verbatim}
(assert (= tile_27_18 6))                             ; endpoint / boundary
(assert (not (hasUp tile_16_1)))                      ; boundary no-exit
(assert (= (hasRight tile_16_2) (hasLeft tile_16_3))) ; adjacency
\end{verbatim}
\end{tcolorbox}

Together, these constraints ensure that all local path segments are
pairwise compatible and that the path interacts with the boundary only
at the prescribed entry and exit. However, they do not prevent the
formation of multiple disconnected components or closed loops, which
necessitates additional global connectivity constraints described
next.

\subsubsection{Global Connectivity Constraints}
\begin{figure}[h!tbp]
  \centering
  \begin{minipage}[b]{0.23\textwidth}
    \includegraphics[width=\textwidth]{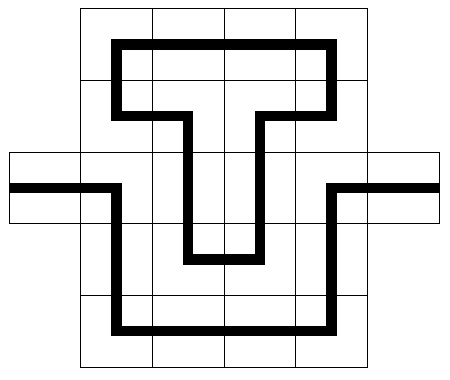}
    \subcaption{}
    \label{fig:na}
  \end{minipage}
  ~
  \begin{minipage}[b]{0.23\textwidth}
    \includegraphics[width=\textwidth]{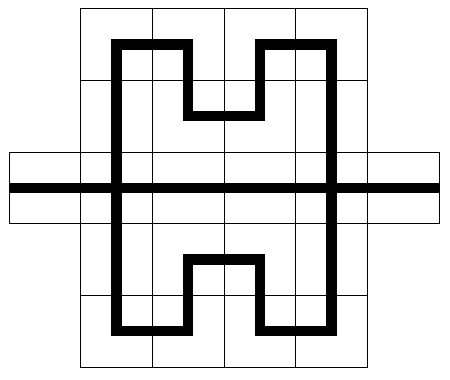}
    \subcaption{}
    \label{fig:nb}
  \end{minipage}
  ~
  \begin{minipage}[b]{0.23\textwidth}
    \includegraphics[width=\textwidth]{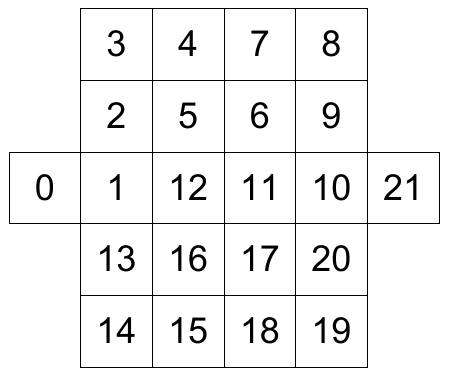}
    \subcaption{}
    \label{fig:nc}
  \end{minipage}
  ~
  \begin{minipage}[b]{0.23\textwidth}
    \includegraphics[width=\textwidth]{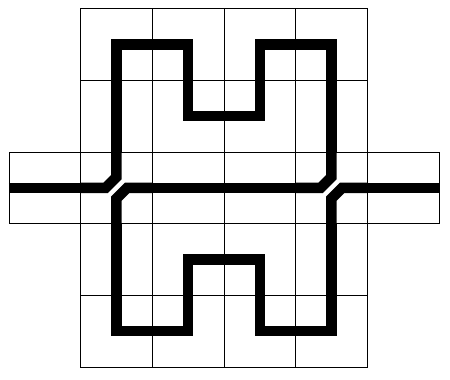}
    \subcaption{}
    \label{fig:nd}
  \end{minipage}
  \caption{Failure modes motivating global connectivity constraints.
    (a) Local consistency alone permits disconnected loops.  (b--d)
    Rank-based connectivity can fail in the presence of crossing
    tiles.}
  \Description{Four small diagrams showing disconnected loops and rank-based connectivity failures involving crossing tiles.}
  \label{fig:abnormal}
\end{figure}

Figure~\ref{fig:na} shows a typical failure mode of purely local
constraints: even when all adjacent tiles are locally compatible, the
selected tiles may form a detached loop rather than a single path
connecting the designated entry and exit. This motivates explicit
global connectivity constraints.

We enforce global connectivity by introducing an explicit predecessor
relation together with a strictly decreasing rank. We write
\(\mathrm{nonWhite}(t_{r,c})=1\) to indicate that the tile at position
\((r,c)\) is not empty and thus participates in the path.

For each grid position \((r,c)\) we introduce an integer rank variable
\(\rho_{r,c}\) and four Boolean parent-choice variables
\(p^\kind{R}_{r,c},p^\kind{L}_{r,c},p^\kind{U}_{r,c},p^\kind{D}_{r,c}\). For example,
\(p^\kind{R}_{r,c}\) means that position \((r,c)\) selects its right
neighbor as its predecessor. We fix the entry position \(s=(r_s,c_s)\)
as the root. The root has rank \(0\) and no parent choice:
\[
\rho_{r_s,c_s}=0,
\qquad
\neg p^\kind{R}_{r_s,c_s}\wedge \neg p^\kind{L}_{r_s,c_s}\wedge
\neg p^\kind{U}_{r_s,c_s}\wedge \neg p^\kind{D}_{r_s,c_s}.
\]
All other non-empty tiles must have positive rank and choose at least
one parent direction:
\begin{equation}
\label{eq:parent}
\mathrm{nonWhite}(t_{r,c}) = 1 \wedge (r,c)\neq (r_s,c_s)
\;\Rightarrow\;
\bigl(p^\kind{R}_{r,c}\vee p^\kind{L}_{r,c}\vee p^\kind{U}_{r,c}\vee p^\kind{D}_{r,c}\bigr)
\;\wedge\; 1 \le \rho_{r,c}\le \vert P\rvert.
\end{equation}

A parent choice must follow an existing local edge to a neighboring
non-empty tile, with matching boundary predicates and strictly
decreasing rank. For instance,
\[
p^\kind{R}_{r,c}\;\Rightarrow\; \rho_{r,c+1}<\rho_{r,c} \wedge
\mathrm{nonWhite}(t_{r,c}) = 1 \wedge \mathrm{nonWhite}(t_{r,c+1})=1
\wedge \mathrm{hasRight}(t_{r,c}) \wedge \mathrm{hasLeft}(t_{r,c+1}) .
\]
The constraints for \(p^\kind{L}_{r,c}, p^\kind{U}_{r,c}, p^\kind{D}_{r,c}\) are
obtained by swapping the neighbor coordinates and the corresponding
directional predicate pairs on the current and adjacent tiles.

These constraints translate directly into SMT-LIB assertions, e.g.:
\begin{tcolorbox}[smtsimple]
\begin{verbatim}
(assert (=> (= (nonWhite tile_13_2) 1)
            (and (>= rank_13_2 1) (<= rank_13_2 507))))
(assert (=> (= (nonWhite tile_13_2) 1)
            (or pR_13_2 pL_13_2 pD_13_2 pU_13_2)))
(assert (=> pR_13_2
            (and (< rank_13_3 rank_13_2)
                 (= (nonWhite tile_13_2) 1)
                 (= (nonWhite tile_13_3) 1)
                 (hasRight tile_13_2)
                 (hasLeft tile_13_3))))
\end{verbatim}
\end{tcolorbox}

Together, these constraints ensure that every participating position
(except the root) selects a locally valid predecessor of smaller rank,
thereby excluding disconnected loops (a directed cycle would force a
strict rank decrease around the cycle).

For simple, non-branching path segments, the combined local adjacency
and global parent--rank constraints behave as expected. However, in
the presence of crossing tiles, the assumption that each grid position
corresponds to a single path position breaks down. Figure~\ref{fig:nb}
shows a representative solution produced under this scheme, which can
satisfy the parent--rank constraints, as illustrated by the rank
assignment in Figure~\ref{fig:nc}.

From a graph-theoretic perspective, this solution admits an Eulerian
single-stroke traversal: only the endpoints have degree \(1\), while
all other tiles have even degree. Such a traversal is shown in
Figure~\ref{fig:nd}. However, the resulting path corresponds to a
purely planar interpretation of the crossing, rather than the intended
woven over--under structure. This motivates a refinement of the
connectivity model at crossings, introduced next.

\subsubsection{Crossing Tiles via Virtual Path Elements}

We address this limitation by modeling a crossing tile as two
co-located virtual path elements: a horizontal element and a vertical
element. Ordinary non-empty, non-crossing tiles contribute a single
base path element.

Formally, we define activation predicates by
\[
\begin{aligned}
A^{\kind{base}}_{r,c}
&= \bigl(\mathrm{nonWhite}(t_{r,c})=1\bigr)\wedge (t_{r,c}\neq 7),
\\
A^{\kind{h}}_{r,c}
&= (t_{r,c}=7),
\qquad
A^{\kind{v}}_{r,c}
= (t_{r,c}=7).
\end{aligned}
\]
Thus a crossing activates two elements \(\kind{h}\) and \(\kind{v}\),
while a non-crossing, non-empty tile activates exactly one element
\(\kind{base}\) in the connectivity model.

For each kind \(k\in\{\kind{base},\kind{h},\kind{v}\}\) we introduce
an integer rank \(\rho^{k}_{r,c}\) and directional parent choice
Booleans
\(p^{\kind{R},k}_{r,c},p^{\kind{L},k}_{r,c},p^{\kind{U},k}_{r,c},p^{\kind{D},k}_{r,c}\). The
global parent--rank constraints introduced in
Equation~\eqref{eq:parent} are now instantiated separately for each
virtual element \(k\), guarded by the corresponding activation
predicate \(A^{k}_{r,c}\). For the split elements \(\kind{h}\) and
\(\kind{v}\), parent choices in unsupported directions are explicitly
disallowed (e.g., \(p^{\kind{R},\kind{v}}_{r,c}\) and
\(p^{\kind{L},\kind{v}}_{r,c}\) are constrained to be false), ensuring that
each element participates in the path only along its admissible
directions. Concretely, for any active non-root element, the
requirement of selecting a parent direction and having a positive rank
is enforced per element:
\[
A^{k}_{r,c}\wedge (r,c)\neq (r_s,c_s)\;\Rightarrow\;
\Bigl(\bigvee_{d\in\{\kind{R},\kind{L},\kind{U},\kind{D}\}} p^{d,k}_{r,c}\Bigr)
\;\wedge\; 1\le \rho^{k}_{r,c}\le |P|.
\]
Since each grid position admits up to three such elements
(\(k\in\{\kind{base},\kind{h},\kind{v}\}\)), this instantiation
results in three parallel instances of the parent--rank constraints at
each position. As in the non-crossing case, at the root
\(s=(r_s,c_s)\) we impose \(\rho^{k}_{r_s,c_s}=0\) whenever
\(A^{k}_{r_s,c_s}\) holds and forbid all parent choices.

Since \(A^{\kind{h}}_{r,c}=A^{\kind{v}}_{r,c}=(t_{r,c}=7)\), a
crossing tile simultaneously activates both virtual elements, and all
subsequent parent--rank constraints apply to \(\kind{h}\) and
\(\kind{v}\) in tandem.

Directional feasibility constraints are extended in the minimal way: a
chosen parent direction must follow an existing local edge, match the
neighboring edge, and decrease rank, but the neighbor rank may belong
either to \(\kind{base}\) (if the neighbor is non-crossing) or to the
corresponding crossing kind (if the neighbor is also a crossing). For
example,
\[
p^{\kind{R},\kind{h}}_{r,c}\;\Rightarrow\;
\bigl(A^{\kind{h}}_{r,c}\wedge \mathrm{hasRight}(t_{r,c})\bigr)
\wedge
\mathrm{hasLeft}(t_{r,c+1})
\wedge
\Bigl(
A^{\kind{base}}_{r,c+1}\wedge \rho^{\kind{base}}_{r,c+1}<\rho^{\kind{h}}_{r,c}
\;\vee\;
A^{\kind{h}}_{r,c+1}\wedge \rho^{\kind{h}}_{r,c+1}<\rho^{\kind{h}}_{r,c}
\Bigr).
\]
In the SMT encoding, this implication expands into a disjunction over
the possible kinds of the neighboring tile; for instance:
\begin{tcolorbox}[smtsimple]
\begin{verbatim}
(assert (=> pR_31_13_h
  (or (and (= tile_31_14 7)
           (= tile_31_13 7) (hasRight tile_31_13) (hasLeft tile_31_14)
           (< rank_31_14_h rank_31_13_h))
      (and (and (not (= tile_31_14 7)) (= (nonWhite tile_31_14) 1))
           (= tile_31_13 7) (hasRight tile_31_13) (hasLeft tile_31_14)
           (< rank_31_14_base rank_31_13_h)))))
\end{verbatim}
\end{tcolorbox}

Analogous implications are added for the other directions, with the
neighbor kind chosen according to the move direction (horizontal uses
\(\kind{base}/\kind{h}\), vertical uses \(\kind{base}/\kind{v}\)).

Note that the coverage definition~\eqref{eq:coverage} remains
unchanged, as it is expressed solely in terms of the tile variables
\(t_{r,c}\) and \(\mathrm{nonWhite}(t_{r,c})\). A crossing still
contributes a single occupied grid position, independent of its
virtual \(\kind{h}/\kind{v}\) split, so the same coverage constraint
(or objective) applies verbatim. When crossing tiles are forbidden,
the corresponding activation predicates are false, and the encoding
degenerates naturally to the planar parent--rank formulation.

The virtual-element encoding describes how crossings are represented
when they occur. In the woven-maze instances used for the constructions
below, we additionally impose a spacing condition: crossing tiles are
not allowed to be adjacent in the grid. In two-dimensional renderings,
this improves visual clarity. In the three-dimensional realization,
the same condition provides the graph-distance separation needed by
the fixed-level height assignment used below.

\subsection{Experiments and Performance}

Using OpenSMT2~\cite{OpenSMT2} on an M1 Max MacBook Pro, we also
recorded the scale and runtime of several representative
instances. For the smallest ``$\infty$'' pattern ($|P|=202$) in
Figure~\ref{fig:inf}, the crossing encoding uses 3232 variables and
5503 constraints and yields a full-coverage solution in about 1.1
seconds. For the larger instances ``A'' ($|P|=447$) and ``rt''
($|P|=421$) in Figure~\ref{fig:examples}, the crossing versions use
7152/12315 and 6736/11585 variables/constraints, respectively, and
yield full-coverage solutions in about 4.8 minutes and 19.1
minutes. The corresponding non-crossing encodings are more
restrictive: at the same target coverage they are often unsatisfiable,
and satisfiability is recovered only after lowering the coverage
bound, sometimes with substantially longer runtimes. In particular,
the non-crossing ``$\infty$'' instance becomes satisfiable only at
195, and the non-crossing ``rt'' instance only at 418; for ``A'', the
non-crossing instance is unsatisfiable at 447 but satisfiable at 441
in about 43.5 minutes. In these examples, allowing crossings permits
higher coverage and makes the instances easier to satisfy. Because the
present OpenSMT2 runs use lower-bound satisfiability checks rather than
a direct maximization objective, optimality for a given instance is
certified by finding a satisfiable bound and proving all larger bounds
unsatisfiable.

\section{Maze Construction}

The SMT encoding described above produces a prescribed solution path:
a sequence of tile choices that traces the desired pattern and
satisfies the planar or layered path constraints. A maze, however,
contains more than its solution path: it also includes additional
passages through the remaining grid positions while preserving the
prescribed entry--exit route as the unique solution.

We separate this construction into two levels. First, the prescribed
solution path is completed to a full topological maze, represented as
a passage graph. This graph is a tree, so the path between the entry
and exit is unique. Second, in the woven case, the completed topology
is realized geometrically by assigning heights to over--under passages
and converting the result into explicit 3D primitives.

\subsection{Maze Completion from a Prescribed Solution Path}

Randomized spanning-tree algorithms are a standard way to generate
perfect mazes, in which every pair of cells is connected by a unique
path; see, for example, Buck's discussion of perfect mazes, Wilson's
algorithm, and randomized Kruskal's algorithm~\cite[Chs.~1, 4,
  and~10]{Buck2015MazesForProgrammers}.  Our setting differs in that
the entry--exit solution path is prescribed in advance. We therefore
treat this path as a fixed subgraph and extend it to a spanning tree
of the completed maze graph. Since the final passage graph is a tree
and already contains the prescribed path from the entry to the exit,
that path remains the unique solution of the completed maze.

In the planar implementation, this completion is performed using a
Wilson-style loop-erased random walk initialized by the prescribed
solution path. The path edges are inserted first, and their grid
positions are marked as already included in the maze. Each remaining
grid position is then attached by a loop-erased random walk that
terminates when it reaches the existing tree. This yields a perfect
maze containing the prescribed solution path.

For woven mazes, the same topological goal is applied to an expanded
over--under graph. A crossing position may contain both an over node
and an under node occupying the same grid position. Required crossings
and solution-path links are inserted first. The remaining candidate
passages are then processed in randomized order and added only when
they merge distinct connected components. This Kruskal-style
completion preserves the tree structure while incorporating the
selected over--under crossings.

For two-dimensional output, the completed passage graph is rendered
using standard wall-and-passage conventions for grid mazes, with inset
corridors used to depict woven passages as in Buck's treatment of
weave rendering \cite[Chs.~2 and~9]{Buck2015MazesForProgrammers}. In
the planar case, walls are drawn between adjacent grid positions that
are not connected by a passage. In the woven case, over- and
under-passages are drawn as separated lanes within a grid position,
making the crossing relation visible without changing the underlying
passage graph.

At this point the maze is complete as a combinatorial object. The
two-dimensional renderings described above are visualizations of this
passage graph, not additional topological constructions. For an
ordinary planar maze, a three-dimensional realization is similarly
direct: the walls of the two-dimensional layout may simply be extruded
to a fixed height. Woven mazes require an additional geometric step,
since their over--under relations must be realized as physical
vertical separation.

\subsection{Three-Dimensional Realization of Woven Mazes}

A woven maze is represented by a passage graph whose crossings record
which local passage goes over and which goes under. To realize this
graph geometrically, we assign integer height levels to the passage
nodes and then convert the resulting height-labeled graph into an
explicit collection of 3D primitives.

The height assignment must satisfy two physical requirements. At each
crossing, the over-passage must be separated from the under-passage by
enough vertical clearance. Along ordinary maze links, adjacent passage
nodes should change height only gradually, so that the transition can
be realized by a short stair. Thus height planning must be solved
globally over the completed woven passage graph.

We separate the three-dimensional realization into height planning and
geometric construction. The height-planning step first states the
physical height constraints, then adopts a stronger fixed-level
convention that keeps all crossings at a small number of integer
levels, and finally proves that this convention is feasible under the
crossing-separation condition used in the construction. Among feasible
assignments, we then choose one that minimizes total height variation
and then total height. The geometric construction step converts the
resulting height-labeled graph into platforms, bridges, connectors,
stairs, and openings.

\subsubsection{Height Constraints and Feasibility}

For the height-planning problem, let \(G=(V,E)\) be the completed
woven maze graph after crossing positions have been expanded into over
and under passage nodes. The over and under nodes at the same crossing
are co-located geometrically, but they are not adjacent in \(G\). Let
\(\mathcal{C}\subseteq V\times V\) be the set of ordered crossing
pairs \((o,u)\), where \(o\) is the over-passage node and \(u\) is the
corresponding under-passage node. We write \(O\subseteq V\) and
\(U\subseteq V\) for the sets of all over-passage and under-passage
crossing nodes.

A height assignment gives each passage node a discrete top-surface
level, initially viewed as a nonnegative integer:
\[
h:V\to \mathbb{Z}_{\ge 0}.
\]
Passage elements have nonzero thickness. Therefore, at a crossing, the
over-passage must be raised enough to provide both empty clearance
above the under-passage and room for the thickness of the over-passage
itself.  Let \(c\) be the required number of empty clearance levels
and \(b\) the number of levels occupied by the over-passage
thickness. We set
\[
g=c+b.
\]
Since \(c\) and \(b\) are positive integers, the smallest
non-degenerate choice is \(c=b=1\), giving \(g=2\). We use this
minimal setting throughout the constructions in this report.

The physical crossing-clearance requirement is
\[
h(o)-h(u)\ge g
\qquad\text{for every }(o,u)\in\mathcal{C}.
\]
The local height-change requirement is
\[
|h(x)-h(y)|\le 1
\qquad\text{for every edge }\{x,y\}\in E.
\]
Together, these conditions say that crossings have enough vertical
clearance and that adjacent passage elements can be connected by local
stairs. By themselves, however, they do not fix the absolute height of
a crossing: different crossings could be realized at different
vertical offsets.

For a simpler and lower-profile geometry, we impose the stronger
fixed-level convention
\[
h(u)=0 \quad (u\in U),
\qquad
h(o)=g \quad (o\in O).
\]
This convention automatically satisfies the crossing-clearance
requirement, but it is not a physical necessity. It is an additional
modeling choice that keeps all under-crossings at level \(0\), all
over-crossings at level \(g\). The only remaining issue is whether
this stronger convention is compatible with the local height-change
constraints.

We now prove that, under the stated crossing-spacing condition, the
fixed-level convention is compatible with the local height-change
constraints for the minimal non-degenerate gap \(g=2\). If there are
no crossings, then the all-zero height assignment is sufficient.
Otherwise \(U\neq\varnothing\), and the completed passage graph \(G\)
is connected, so every node has finite distance to \(U\). Define
\(d_G(v,U)\) to be the shortest graph distance from \(v\) to the set
\(U\), and assign heights by
\[
h(v)=\min\{2,d_G(v,U)\}.
\]

We first verify the fixed crossing heights. For every under-passage
node \(u\in U\), we have \(d_G(u,U)=0\), hence \(h(u)=0\). For an
over-passage node \(o\in O\), we need to show that \(h(o)=2\). The
expanded over and under nodes at the same crossing are co-located
geometrically but are not adjacent in \(G\), and \(o\notin U\), so
this crossing does not put \(o\) at distance \(0\) or \(1\) from
\(U\). The remaining way for \(o\) to be at distance \(1\) from \(U\)
would be an edge in \(G\) from an over-passage crossing node to an
under-passage crossing node at a neighboring crossing. By the spacing
condition stated above, the completed woven-maze construction avoids
adjacent crossings, and in the expanded passage graph this prevents
such over-to-under crossing
adjacencies. Therefore \(d_G(o,U)\ge 2\), so \(h(o)=2=g\). Thus the
fixed crossing heights are satisfied.

It remains to check the local height-change constraints. If
\(\{x,y\}\in E\), then the distance-to-\(U\) function changes by at
most one across this edge:
\[
|d_G(x,U)-d_G(y,U)|\le 1.
\]
Clamping these distances by \(\min\{2,\cdot\}\) cannot increase this
difference, so
\[
|h(x)-h(y)|\le 1.
\]
Therefore the fixed-level convention is compatible with the local
height-change constraints, and the constructed height assignment is
feasible.

For larger choices of \(g\), the same distance-based construction
would require the stronger graph-distance separation
\(d_G(o,U)\ge g\) for every \(o\in O\).

\subsubsection{Optimal Height Assignment}

The distance-based construction above proves that, for the fixed-level
convention with \(g=2\) used here, feasible height assignments exist
under the stated crossing-spacing condition. We now optimize over the
same feasible set: the under-passage crossing nodes remain fixed at
level \(0\), the over-passage crossing nodes remain fixed at level
\(g\), and every maze edge must still satisfy the local height-change
constraint.

Although heights were introduced as nonnegative integers, no optimum
is lost by restricting the free nodes to the finite range
\(\{0,1,\ldots,g\}\). Indeed, given any feasible assignment \(h\),
define a new assignment \(h'(v)=\min\{h(v),g\}\) for every node
\(v\). This simultaneous clamping preserves the fixed crossing
heights, cannot increase any edge height difference, and weakly
decreases the total height. Thus we use the height domain
\[
D_v =
\begin{cases}
\{0\}, & v\in U,\\
\{g\}, & v\in O,\\
\{0,1,\ldots,g\}, & \text{otherwise}.
\end{cases}
\]

Among all assignments satisfying these domains and the local
height-change constraint, we minimize the lexicographic objective
\[
\left(
\sum_{\{x,y\}\in E}|h(x)-h(y)|,\;
\sum_{v\in V}h(v)
\right).
\]
The first component minimizes the total height variation across maze
links; under the constraint \(|h(x)-h(y)|\le 1\), this is equivalently
the number of links that require a stair. The second component breaks
ties by preferring a lower overall height profile.

Since the completed maze graph \(G\) is a tree, the optimization can
be solved by dynamic programming. Root \(G\) arbitrarily. For each
node \(v\), let \(T_v\) be the subtree rooted at \(v\). Once the
height of \(v\) is fixed, the choices in the child subtrees interact
only through the height differences on the edges from \(v\) to its
children.

For each \(z\in D_v\), define \(F_v(z)\) to be the best lexicographic
cost of the subtree \(T_v\), assuming \(h(v)=z\). If \(v\) is a leaf,
then
\[
F_v(z)=(0,z).
\]
For an internal node \(v\), the cost is obtained by combining the
optimal choices for its children:
\[
F_v(z)
=
(0,z)
+
\sum_{w\in \mathrm{children}(v)}
\min_{\substack{z'\in D_w\\ |z-z'|\le 1}}
\left(
F_w(z') + (|z-z'|,0)
\right).
\]
Here addition of pairs is componentwise, and all minima are taken in
lexicographic order. If no child height \(z'\in D_w\) satisfies
\(|z-z'|\le 1\), then the state \(F_v(z)\) is infeasible.

After all subtree tables are computed bottom-up, the root height is
chosen by the same lexicographic rule. The optimal height assignment
is then recovered by backtracking the child-height choices recorded in
the recurrence. A direct implementation checks all pairs of parent and
child heights and has running time \(O(|E|g^2)\). Under the minimal
fixed-level setting used here, \(g=2\), so the running time is linear
in the size of the maze graph.

\subsubsection{Geometric Construction}

Once the height plan has been computed, the height-labeled maze graph
is converted into a set of three-dimensional primitives. This
conversion is separated from any particular rendering or export
backend: the intermediate 3D plan records the geometric roles of maze
components without depending on a specific output format.

The construction begins with a rectangular base plate, when one is
requested, and then places geometric primitives for the completed
passage graph. Ordinary passage nodes, including under-passage
crossing nodes, become platforms at their assigned height levels. The
over node of a crossing is represented instead as a bridge element
suspended above the corresponding under-passage. Thus the same
topological crossing information used in the height plan is realized
geometrically as a local overpass-and-underpass structure.

Edges of the passage graph are converted according to the height
difference between their endpoints. Adjacent nodes of equal height are
connected by horizontal connector elements. Adjacent nodes whose
heights differ by one level are connected by a short sequence of stair
segments, so that vertical changes are spread across the link rather
than occurring as a discontinuous jump. The designated entry and exit
are represented by opening extensions from the appropriate boundary
nodes of the maze.

Raised parts of the maze may also require supports. The construction
therefore allows either solid support below platforms or selected
support posts beneath raised elements. These supports are geometric
additions only: they do not change the passage graph, the solution
path, or the connectivity of the maze.

\begin{figure}[htbp]
  \centering
  \includegraphics[width=0.62\linewidth]{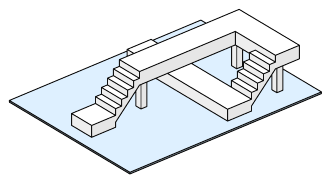}
  \caption{Local geometric construction of a woven crossing.}
  \Description{A stylized local three-dimensional crossing construction with a lower passage, raised bridge, platforms, stairs, support posts, and a base plate.}
  \label{fig:crossing-construction}
\end{figure}

Figure~\ref{fig:crossing-construction} gives a compact local view of
these geometric primitives. The lower passage realizes the
under-crossing, the raised bridge realizes the over-crossing, and the
nearby platforms, stairs, and support posts show how neighboring
passages and auxiliary support geometry are attached.

When the solution path is displayed, it is added as a raised ribbon
following the unique entry--exit path in the completed maze tree. This
ribbon is treated as a separate visual primitive rather than as part
of the walkable maze topology. Its width, lift above the passage
surface, and thickness are chosen so that it remains inside the
passage footprint while staying below the crossing clearance.

Physical dimensions are controlled by a small set of parameters,
including the cell size, passage width, passage-body thickness,
vertical level height, stair subdivision count, base-plate thickness,
support style, and solution-ribbon dimensions. If \(z_0\) is the
physical top height of level \(0\), and \(\Delta z\) is the physical
height represented by one integer level, then the top-surface height
of a node at integer level \(\ell\) is
\[
z(\ell)=z_0+\ell\Delta z.
\]
The discrete height plan therefore determines the vertical placement
of all platforms, bridges, connectors, stairs, openings, and
solution-ribbon segments.

Finally, the construction checks that the physical dimensions are
compatible with the discrete height gap. In particular, the vertical
clearance below an over-crossing must meet the required empty
clearance after accounting for the thickness of the passage element.
If the solution ribbon is included, its lift above the passage surface
must also remain below this available clearance. Thus the discrete
height assignment supplies the combinatorial structure, while the
geometric construction turns it into an explicit three-dimensional
maze with walkable passages, crossings, supports, openings, and an
optional highlighted solution.

\section{Examples}

\begin{figure}[h!tbp]
  \centering
  \includegraphics[width=0.96\linewidth]{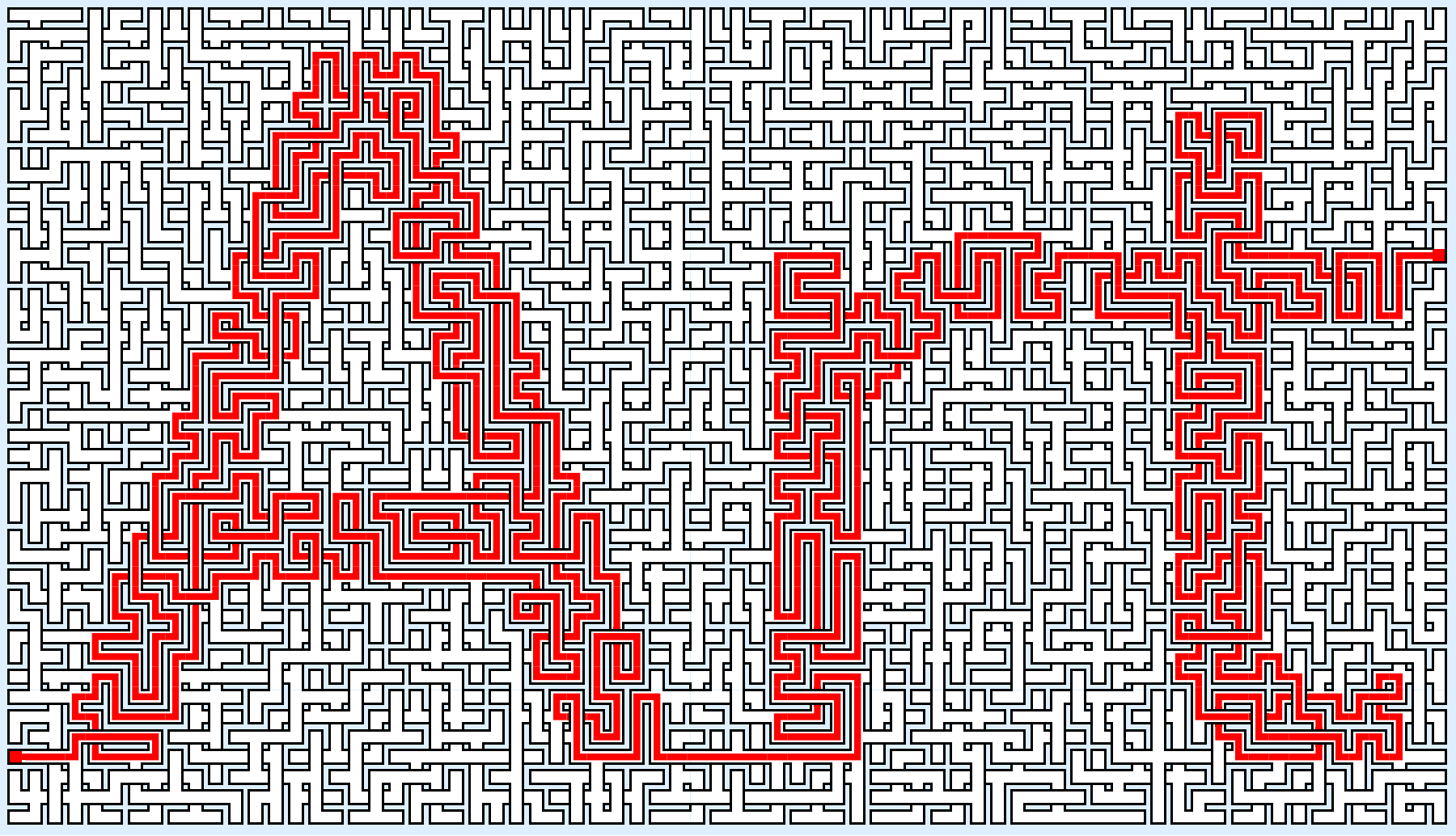}
  \caption{A large woven maze generated by the proposed encoding.}
  \Description{A large woven maze whose highlighted solution path forms the word Art.}
  \label{fig:examples}
\end{figure}

Figure~\ref{fig:examples} shows a large woven maze produced by the
proposed encoding, with the solution path highlighted in red. Crossing
tiles enable a single-stroke path that passes through intersections in
an over--under fashion. For path construction, the underlying ``Art''
pattern was split into two subpatterns, ``A'' and ``rt'', which were
solved separately and then combined before the final maze-generation
step. To form a connected grid, additional links are manually
introduced between otherwise disjoint letter strokes and at the
designated endpoints.

\begin{figure}[htbp]
  \centering
  \includegraphics[width=0.98\linewidth]{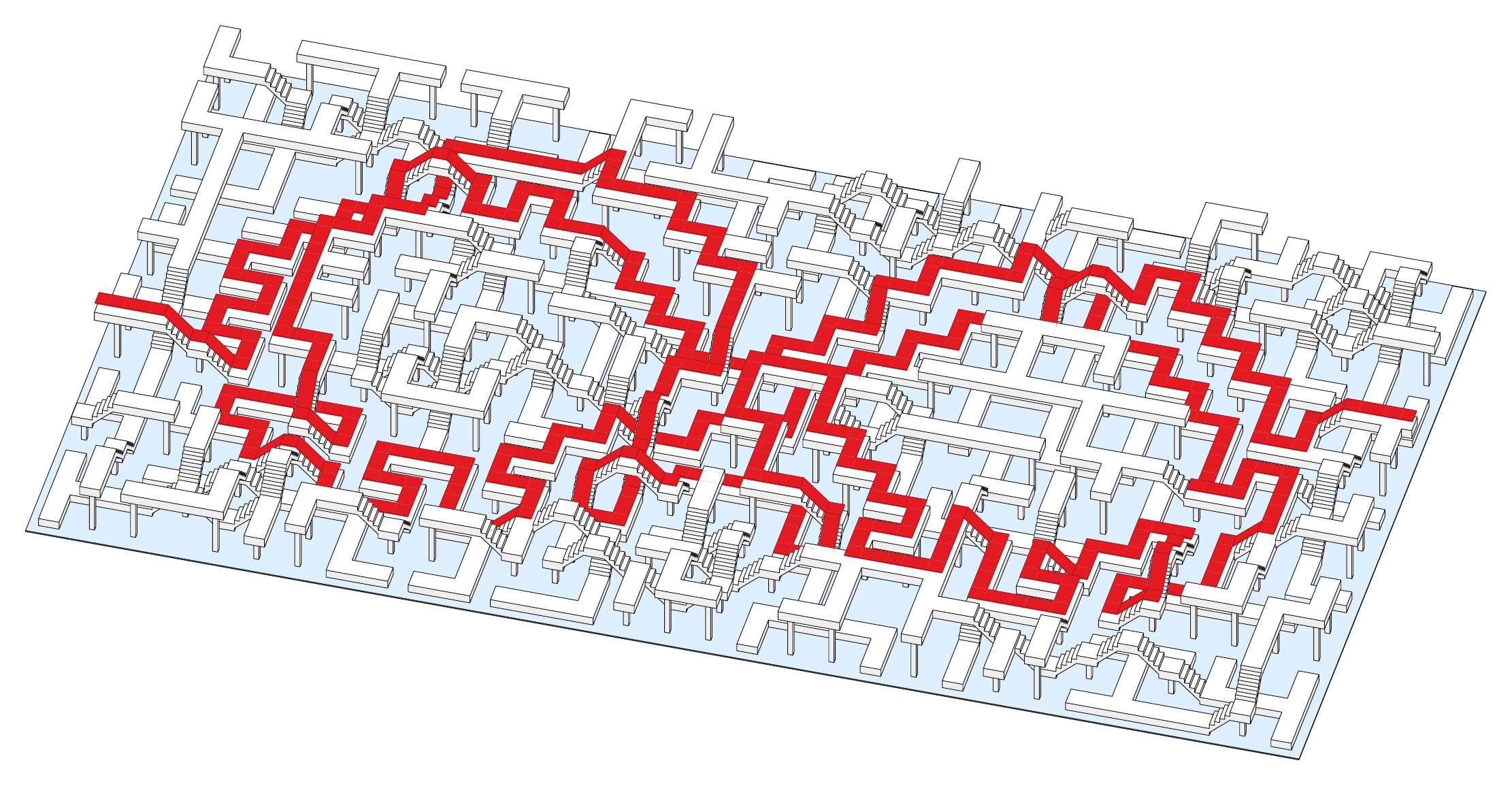}
  \caption{Orthographic three-dimensional rendering of the woven maze
    realization for the ``$\infty$'' pattern.}
  \Description{An orthographic three-dimensional rendering of a woven
    maze for an infinity-shaped solution path, with light-colored
    passages and a red highlighted solution ribbon.}
  \label{fig:inf-3d}
\end{figure}

Figure~\ref{fig:inf-3d} shows an orthographic view of a
three-dimensional woven maze realization produced by the construction
described above. The light-colored geometry is the completed
three-dimensional maze, and the red ribbon marks the unique
entry--exit solution path for the ``$\infty$'' pattern.

\section{Discussion and Conclusions}

We presented an SMT-based pipeline for constructing maze structures
from prescribed rasterized patterns. The central path-synthesis step
combines local tile-connectivity rules with global parent--rank
constraints, and supports both planar paths and layered woven paths
within a unified encoding. The resulting solution path is then
completed into a perfect maze by standard spanning-tree
maze-generation methods initialized with the prescribed path. For
woven mazes, we further give a height assignment and geometric
construction that realize the over--under crossings as explicit
three-dimensional structure.  Together, these steps show how
constraint-based reasoning can connect visual pattern design, maze
topology, and geometric realization in a single computational
pipeline.

\section*{Code Availability}

The implementation is split across two public GitHub repositories. The
SMT-based path synthesis code, including the Lean implementation,
SMT-LIB encoder, solver interface, and manuscript sources, is
available at \url{https://github.com/txyyss/opt-maze}. The Python code
for completing synthesized paths into planar and woven mazes,
assigning heights for three-dimensional realizations, exporting
geometry, and rendering examples is available at
\url{https://github.com/txyyss/maze-construction}. The two
repositories correspond to the two main computational stages of this
report: path synthesis and maze construction.

\begin{acks}
AI-assisted tools, including ChatGPT, were used for language editing,
exploratory technical discussion, assistance with parts of the
implementation, and for first suggesting the fixed-level existence
proof and optimal-height dynamic-programming algorithm.
All mathematical claims, algorithms, and experimental results were
independently examined and validated by the author.
\end{acks}


\bibliographystyle{ACM-Reference-Format}
\bibliography{maze}

\end{document}